\documentclass[runningheads,a4paper]{llncs}
\usepackage{amssymb}
\usepackage{graphicx}
\usepackage{subfig}
\usepackage[hidelinks]{hyperref}
\usepackage{url}
\usepackage{comment}
\usepackage{booktabs}
\usepackage[colorinlistoftodos]{todonotes}

\urldef{\mailsa}\path|{sebastian.stabinger, antonio.rodriguez-sanchez, justus.piater}@uibk.ac.at|
\newcommand{\keywords}[1]{\par\addvspace\baselineskip
  \noindent\keywordname\enspace\ignorespaces#1}

\newcommand{\samples}[1]{
  \def\imgsize{0.10}
  \begin{figure}[htbp]
    \centering
    \subfloat[Class 1]
    {\fbox{\includegraphics[width=\imgsize\textwidth]{images/dataset_#1/sample_0_0000.png}}
      \fbox{\includegraphics[width=\imgsize\textwidth]{images/dataset_#1/sample_0_0001.png}}}
    \hspace{2mm}
    \subfloat[Class 2]
    {\fbox{\includegraphics[width=\imgsize\textwidth]{images/dataset_#1/sample_1_0000.png}}
      \fbox{\includegraphics[width=\imgsize\textwidth]{images/dataset_#1/sample_1_0001.png}}}
    \hfill
    \caption{Example images for Problem #1}
    \label{fig:prob#1_example}
  \end{figure}
}

\begin{document}

\setlength{\tabcolsep}{1mm}

\newcommand{\citez}[1]{\cite{#1}}

\renewcommand{\floatpagefraction}{.8}%

\newcommand{\squeezeup}{\vspace{-15mm}}
\newcommand{\littlesqueezeup}{\vspace{-5mm}}

\mainmatter  

\title{25 years of CNNs: Can we compare to human abstraction
  capabilities?}

\author{Sebastian Stabinger \and
  Antonio Rodr\'iguez-S\'anchez \and
  Justus Piater}

\institute{University of Innsbruck,\\
  Technikerstrasse 21a, 6020 Innsbruck, Austria\\
  \mailsa\\
  \url{http://iis.uibk.ac.at}}

\maketitle

\begin{abstract}
  We try to determine the progress made by convolutional neural
  networks over the past 25 years in classifying images into abstract
  classes. For this purpose we compare the performance of LeNet to
  that of GoogLeNet at classifying randomly generated images which are
  differentiated by an abstract property (e.g., one class contains two
  objects of the same size, the other class two objects of different
  sizes). Our results show that there is still work to do in order to
  solve vision problems humans are able to solve without much
  difficulty.

\end{abstract}

\textbf{To appear in the proceedings of ICANN 2016, Springer}

\section{Introduction}
Deep learning methods have gained interest from the machine learning
and computer vision research communities over the past several years
because these methods provide exceptional performance for a vast
majority of classification tasks. An important example of deep
learning methods are Convolutional Neural Networks (CNNs) --- first
introduced in 1989 by LeCun {\it et al.}
\citez{lecun1989backpropagation} --- which have become popular for
object classification. CNNs were more widely used after the deep CNN
from Krizhevsky {\it et al.} \citez{krizhevsky2012imagenet}
outperformed state-of-the-art methods by a wide margin in the
``ImageNet Large Scale Visual Recognition Competition'' of 2012.

Convolutional neural networks consist of multiple layers of nodes,
also called neurons. One important layer type is the convolutional
layer, from which the networks obtain their name. In a convolutional
layer, the responses of the nodes depend on the convolution of a
region of the input image with a kernel. Additional layers introduce
non-linearities, rectification, pooling, etc. The goal of training a
CNN lies in optimizing the network weights (including the kernels used
for convolution) using image-label pairs to best reconstruct the
correct label, given an image. During testing, the network is
confronted with novel images and expected to generate the correct
label. The network is trained by gradient descent which is calculated
by backpropagation of labeling errors. The general idea of CNNs is to
automatically learn the features needed to distinguish classes and
generate increasingly abstract features as the information is
transferred to higher layers.

Since CNNs are very popular at the moment and are being perceived ---
in parts of the computer vision community --- as achieving human-like
performance, we wanted to test their applicability on visual tasks
slightly outside the mainstream which are still trivially solved by
humans.

\section{Materials and Methods}
\label{sec:materials-methods}

\subsection{The Dataset}
We use the framework presented by Fleuret {\it et al.}
\citez{fleuret2011comparing} to generate our dataset consisting of 23
different problems which are briefly summarized as follows: Each
problem consists of two classes of images. Images of the first class
exhibit some abstract property which is not present in images of the
second class and vice versa. \autoref{fig:prob1_example} shows
examples of the two classes for problem one. Both classes contain two
random objects. In the first class the objects are different, while in
the second they are identical. The goal is to assign the correct class
to previously unseen images.
These problems are reminiscent of the Bongard problems presented by
Bongard \citez{bongard1970} and further popularized by Hofstadter
\citez{douglas1979godel}.

\littlesqueezeup{}
\samples{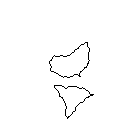}
\littlesqueezeup{}

For each class of each problem we generate $20000$ training images. We
also generate an additional $10000$ images per class and problem as a
testing set. The size of the generated images varies depending on the
used CNN. We chose images of $64 \times 64$ pixels for LeNet, and
$224 \times 224$ pixels for GoogLeNet.

\subsection{Learning Framework}
For training the CNNs, we used Caffe by Jia {\it et al.}
\citez{jia2014caffe}. More specifically, we used the implementations
of LeNet and GoogLeNet provided with Caffe. Only slight adaptations
were made to some hyperparameters. See the appendix for concrete
values. In addition, we used ADAM by Kingma \& Ba
\citez{kingma2014adam} as the solver method instead of stochastic
gradient descent and changed the last fully connected layer to only
contain two neurons representing our two classes.

\section{Experimental Evaluation}
Since we want to know how much progress has been made between the
first CNNs and a state-of-the-art model, we compare the performance of
LeNet by LeCun {\it et al.} \citez{lecun1989backpropagation} from 1989
to GoogLeNet by Szegedy {\it et al.} \citez{szegedy2014going} from
2014. We chose GoogLeNet as the modern CNN since it is a very popular
architecture and it performed best in a number of categories in
ILSVRC14. LeNet was chosen since it is the oldest widely known CNN.

We train one instance of LeNet and GoogLeNet for each problem using
$20000$ training images per class. The trained networks are then
evaluated on a testing set containing $10000$ previously unseen images
per class for the same problem. The reported accuracy of the network
is the proportion of correctly classified images to the number of all
tested images.
For three problems (3, 11, 13) from Fleuret {\it et al.}
\cite{fleuret2011comparing} we could not generate images of the
correct size. Since we do not think it will influence the overall
conclusion, we excluded those problems from our evaluation.

\section{Results}
\label{sec:results}
\autoref{tab:results} gives an overview of the achieved accuracy of
both tested network architectures, the method presented by Fleuret
{\it et al.} \citez{fleuret2011comparing}, and human test subjects. In
addition, the table gives a short description of the properties which
are used to differentiate the two classes.
{\def\arraystretch{0.5}
  \begin{table}[htb]
    \centering
    \caption{Accuracy comparison of presented methods. The two groupings
      consist of problems which either need shape comparison to be
      solved or not. Accuracy of LeNet and GoogLeNet are experimentally
      determined in this paper. Fleuret are results from the best
      performing system proposed by Fleuret {\it et al.}
      \citez{fleuret2011comparing} (Boosting with feature group 3). The
      human results are estimated accuracies of participants also tested
      by Fleuret {\it et al.} \citez{fleuret2011comparing} and
      reinterpreted for this paper.}
    \begin{tabular}{rrrrrrrl}
      \toprule
      Problem & LeNet & GoogLeNet & Fleuret & Human & Difference between Classes\\
      \midrule
      1 & 0.57 & 0.50 & 0.98 & 0.98 & Compare\\
      5 & 0.54 & 0.50 & 0.87 & 0.90 & Compare \& grouping\\
      6 & 0.76 & 0.86 & 0.76 & 0.70 & Compare \& grouping\\
      7 & 0.53 & 0.50 & 0.76 & 0.90 & Compare \& grouping\\
      8 & 0.94 & 0.91 & 0.90 & 1.00 & Compare \& relative position\\
      15 & 0.52 & 0.50 & 1.00 & 0.95 & Compare\\
      16 & 0.98 & 0.50 & 1.00 & 0.78 & Compare\\
      17 & 0.75 & 0.95 & 0.67 & 0.78 & Compare \& relative position\\
      19 & 0.51 & 0.50 & 0.61 & 0.98 & Compare\\
      20 & 0.55 & 0.50 & 0.70 & 0.98 & Compare\\
      21 & 0.51 & 0.51 & 0.50 & 0.83 & Compare\\
      22 & 0.59 & 0.50 & 0.97 & 1.00 & Compare\\
      \midrule
      2 & 1.00 & 1.00 & 0.98 & 1.00 & Relative position\\
      4 & 0.98 & 1.00 & 0.93 & 1.00 & Relative position\\
      9 & 0.93 & 1.00 & 0.68 & 0.93 & Size \& relative position\\
      10 & 0.99 & 1.00 & 0.94 & 0.98 & Relative position\\
      12 & 0.97 & 1.00 & 0.84 & 0.95 & Size \& relative position\\
      14 & 0.90 & 1.00 & 0.73 & 0.98 & Alignment\\
      18 & 0.99 & 0.99 & 0.99 & 0.93 & Grouping\\
      23 & 0.87 & 1.00 & 0.75 & 1.00 & Relative position\\
      \midrule
      Average & 0.77 & 0.76 & 0.83 & 0.93 & \\
      \bottomrule
    \end{tabular}
    \label{tab:results}
  \end{table}}

At first glance, CNNs do not seem to have made much progress over the
last 25 years with the types of problems we tested, and even compare
very unfavorably to the boosting method presented by Fleuret {\it et
  al.} \cite{fleuret2011comparing}. The average accuracy of GoogLeNet
even decreased slightly compared to LeNet.

Upon closer inspection, there seem to be two groups of problems: Ones
which require the comparison of shapes and ones that do not. If we
only consider problems which do not, the two CNNs perform very well.
LeNet has an average accuracy of $0.95$ and GoogLeNet achieves
practically perfect accuracy. Both also compare very favorably to the
method presented by Fleuret {\it et al.} \citez{fleuret2011comparing}
which achieves a mean accuracy of $0.86$ on this subset of problems.
We will discuss those two subsets of problems in the following
sections in more detail.

\subsection{Problems not involving Comparisons}
\label{sec:relative-position}
Problems 2, 4, 9, 10, 12, 14, 18, and 23 can be differentiated by the
relative positioning or grouping of the shapes. The shapes themselves
are not relevant to the classification except for problems 9 and 12,
where the size of some of the shapes play a roll in the
classification. Apparently, those problems can be solved by detecting
local and global features alone. Hence CNNs work well on those
problems.

\subsection{Problems involving Comparisons}
\label{sec:compare}
Problems 1, 5, 6, 7, 8, 15, 17, 19, 20, 21, and 22 involve comparing
shapes in one way or another. To solve these problems, an agent
has to be able to decide whether two shapes are similar or not at one
stage of the classification process; e.g., in problem 1
(\autoref{fig:prob1_example}) the two classes only differ by whether the
two presented shapes are identical or not.
Except for problems 6, 8, 16, and 17, LeNet as well as GoogLeNet do
not achieve accuracies significantly above chance.

Problems 6, 8, and 17 seem to be solvable by the tested CNNs although
they in theory also require the comparison of shapes. \emph{Problem 6}
(\autoref{fig:prob6_example}) presents two pairs of identical shapes
and the two classes are separated by whether the distances between
each pair is the same or not. \emph{Problem 8}
(\autoref{fig:prob8_example}) presents two shapes of differing size.
One class always contains a small shape inside a bigger version of the
same shape. The other class either has a smaller shape inside a
different, bigger shape or two identical shapes which are not nested.
\emph{Problem 17} (\autoref{fig:prob17_example}) presents four shapes,
of which three are identical. The two classes are separated by whether
the distance between the identical shapes are all the same or not.

\samples{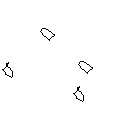}
\samples{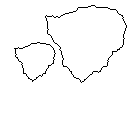}
\samples{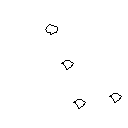}

In theory, an agent has to be able to compare shapes to solve problems
6, 8, and 17; otherwise the additional information, like relative
position, does not matter. We had the suspicion that the generation
process for these problems imparts some unwanted pattern to the images
which the CNNs can use to separate the classes thus avoiding the need
to compare shapes. If this is the case, we can expect the same
accuracy even if images of both classes contain identical shapes.
Theoretically this should mean that those modified problems are not
solvable. Training and testing the CNNs with those modified problems
gives us similar results (\autoref{tab:identresults}) to the original
problems, which indicates that the CNNs are exploiting some unintended
pattern in the data and comparing the shapes does not contribute to
the classification.

\begin{table}[htb]
  \centering
  \caption{Results for problems 6, 8, and 17 when all images only
    contain identical shapes.}
  \begin{tabular}{rrrl}
    \toprule
    Problem & LeNet & GoogLeNet & Difference Between Classes\\
    \midrule
    6 & 0.75 & 0.85 & Compare and grouping\\
    8 & 0.95 & 0.90 & Compare and relative position\\
    17 & 0.77 & 0.93 & Compare and relative position\\
    \bottomrule
  \end{tabular}
  \label{tab:identresults}
\end{table}

\emph{Problem 16} (\autoref{fig:prob16_example}) requires the agent to
decide whether shapes on the right side are identical copies of the
shapes on the left, or whether they are vertically mirrored.
Surprisingly, LeNet solves this problem almost perfectly, with an
accuracy of $0.98$, while GoogLeNet cannot solve it at all, with an
accuracy of $0.5$. We suspected this to be an artifact and that
generating the images with a relatively small size of $64 \times 64$
pixels for LeNet adds some unwanted pattern to the images which the
network can exploit. Since GoogLeNet uses images with a size of
$224 \times 224$ pixels it would not profit from this. To test this
hypothesis, we trained LeNet using images with a size of
$128 \times 128$ pixels, and, as expected, the accuracy dropped to
$0.5$. \samples{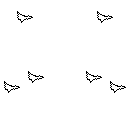}
\littlesqueezeup{}

\subsection{Human Performance}
\label{subsec:human-performance}
Fleuret {\it et al.} \citez{fleuret2011comparing} presented
experiments to determine the performance of humans on the same dataset
we use for our experiments.
Each participant was tested on all problems. For each of the problems,
an example which is randomly chosen from one of the two classes is
presented and the participant has to indicate whether it is from class
one or two. After choosing a class, the correct answer is revealed and
the next example is shown. All previously seen images are kept on the
screen with their correct class. What is recorded in the experiment is
the number of examples the person has to see until he or she
consistently chooses the correct class. It is also recorded if a test
subject can not solve a problem at all.

Unfortunately, the mode of testing is sufficiently different from the
way machine learning solutions are evaluated that a direct
quantitative comparison is difficult. To get some accuracy values we
can compare other methods to we define accuracy of humans as follows.
We assume a person which was able to solve a problem to have an
accuracy of $1.0$ and one which was not of $0.5$. We can then
calculate an expected accuracy of the whole group of test subjects with
\begin{equation}
  \label{eq:2}
  a=\frac{p_a+\frac{p_n}{2}}{n}
\end{equation}
where $a$ is the accuracy, $p_a$ is the number of participants who
were able to solve this problem, $p_n$ being the number of
participants who were not able to solve the problem and $n$ being the
number of all participants. The accuracies reported in
\autoref{tab:results} were calculated from the original data reported
by Fleuret {\it et al.} \citez{fleuret2011comparing} using
\autoref{eq:2}.

\section{Discussion}
Looking at the results of our experiments one can come to very
different conclusions. Simply looking at the overall performance looks
very disappointing. The over 25 year old LeNet is better than the
current GoogLeNet, although only marginally. A closer inspection
reveals that there is a problem class which neither of the CNNs is
capable of solving at all; namely problems which require the
comparison of shapes. We showed that the few problems in this class
which the CNNs can learn are actually learned because of some
unexpected side effects of image generation. We conclude that CNNs
have an inherent problem when it comes to comparative features. It
should be noted that neither humans nor the boosting method employed
by \cite{fleuret2011comparing} show this big performance gap between
the two subsets. The mean accuracy of the boosting method is $0.81$
for problems with shape comparison versus $0.86$ for problems without.
The human test subjects show a mean accuracy of $0.90$ and $0.97$ for
the two subsets respectively.

If we accept that CNNs are generally not capable of solving problems
containing shape comparison, the results look a lot better. Not only
do both networks perform very well on the other problems, but
GoogLeNet achieves, for all intents and purposes, perfect accuracy. It
even outperforms the human test subjects. Obviously, the CNNs need a
much larger training set to achieve those accuracies. Where human
subjects usually need below 20 images and often only require 2 images
to correctly learn the class and achieve perfect accuracy, GoogLeNet
generally needs about 4000 images to achieve an accuracy $\geq 0.99$
(problem~2:~400~images, problem~4:~4000, 9:~4000, 10:~4000, 12:~40000,
14:~40000, 18:~4000, 23:~4000). Of course, humans have a lot of prior
knowledge, so the results are hard to compare. An interesting
difference between machine learning algorithms and humans is the fact
that an algorithm can have an accuracy of e.g. 80\% on these abstract
problems, but human subjects generally either understand what
separates the two classes and achieve an accuracy of 100\%, or do not
understand it and have an accuracy close to pure chance. This suggests
that the underlying principles of classification are probably very
different.

Further, our experiments show how difficult it can be to evaluate CNNs
on abstract problems. One has to be extremely careful to guarantee
that the network is actually solving the problem one wants to test and
does not use some additional superficial pattern. In our case it would
have appeared as if CNNs can in fact compare shapes because they were
able to solve problems 6, 8, and 17 quite successfully. Only close
scrutiny revealed that the networks were in fact exploiting patterns
which were a side effect of the dataset generation.

We think it will be useful to further investigate the performance of
deep learning methods on more abstract problems than are usually
considered since it can reveal a lot about the shortcomings and
strengths of specific methods and might inform further advances of the
methods. We further hypothesize that if the shape comparison problem
of CNNs can be solved they would presumably also perform better on
more common tasks.

\bibliography{typeinst}

\section{Appendix}
\label{sec:appendix}

\begin{itemize}
\item Parameters used for LeNet: iterations~=~$25000$, base learning
  rate~=~$0.001$, weight decay~=~$0.00005$, solver~=~ADAM,
  $\beta_1=0.9$, $\beta_1=0.999$, $\epsilon=10^{-8}$.
\item Parameters used for GoogLeNet: iterations~=~$25000$, base learning
  rate~=~$0.001$, average loss~=~100, weight decay~=~$0.002$,
  solver~=~ADAM, $\beta_1=0.9$, $\beta_1=0.999$, $\epsilon=10^{-8}$.
\end{itemize}

\section{Acknowledgments}
We want to thank nVidia for supporting this research with their
``NVIDIA Hardware Grant''. We also want to thank François Fleuret for
providing us with the dataset used in this paper.

\end{document}